\pdfoutput=1

\documentclass[11pt]{article}

\usepackage[]{acl}

\usepackage{times}
\usepackage{latexsym}

\usepackage[T1]{fontenc}
\usepackage[utf8]{inputenc}

\usepackage{microtype}
\usepackage{multirow}
\usepackage{graphicx}
\usepackage{booktabs}
\usepackage{tikz}
\usepackage{amssymb}
\usepackage{textcomp}
\usepackage{colortbl}
\usepackage{bibentry}
\usepackage{ulem}
\usepackage{float}
\usepackage{arydshln}
\usepackage{threeparttable}
\usepackage{mathtools}
\usepackage{bbm}
\usepackage{chngpage}
\usepackage{array}
\usepackage{adjustbox}
\usepackage{hyperref}

\usepackage{soul}
\usepackage{color}

\colorlet{soulred}{red!30}

\colorlet{soulbleu}{cyan!20}

\colorlet{soulgreen}{green!20}
\DeclareRobustCommand{\hlgreen}[1]{{\sethlcolor{soulgreen}\hl{#1}}}

\colorlet{soulyellow}{yellow!40}

\colorlet{soulorange}{orange!50}

\colorlet{soulpurple}{blue!50}

\definecolor{cornflowerblue}{rgb}{0.39, 0.58, 0.93}
\colorlet{soulcornflowerblue}{cornflowerblue!50}

\makeatletter

\def\uwave{\bgroup \markoverwith{\lower3.5\p@\hbox{\sixly \textcolor{red}{\char58}}}\ULon}

\def\mask{{\sc[mask]}}

\title{Read before Generate! Faithful Long Form Question Answering with Machine Reading}

\author{Dan Su$^1$\thanks{$^*$ Work done during an internship at Huawei Noah's Ark lab}, Xiaoguang Li$^2$, Jindi Zhang$^2$, Lifeng Shang$^2$, Xin Jiang$^2$, Qun Liu$^2$, Pascale Fung$^1$ \\
$^1$Center for Artificial Intelligence Research (CAiRE)\protect\\
  The Hong Kong University of Science and Technology, Clear Water Bay, Hong Kong\\
$^2$Huawei Noah’s Ark Lab \\
\texttt{dsu@connect.ust.hk}, \texttt{lixiaoguang11@huawei.com}
}

\date{}

\begin{document}
\maketitle
\begin{abstract}
Long-form question answering (LFQA) aims to generate a paragraph-length answer for a given question. While current work on LFQA using large pre-trained model for generation are effective at producing fluent and somewhat relevant content, one primary challenge lies in how to generate a faithful answer that has less hallucinated content. We propose a new end-to-end framework that jointly models answer generation and machine reading. The key idea is to augment the generation model with fine-grained, answer-related salient information which can be viewed as an emphasis on faithful facts. State-of-the-art results on two LFQA datasets, ELI5 and MS MARCO, demonstrate the effectiveness of our method, in comparison with strong baselines on automatic and human evaluation metrics. A detailed analysis further proves the competency of our methods in generating fluent, relevant, and more faithful answers.


\end{abstract}

\section{Introduction}

Long-form question answering (LFQA) is a task to generate an in-depth, paragraph-length answer for a given question~\cite{fan2019eli5}. It is important since many of the everyday questions that humans deal with and pose to search engines require multi-sentence explanations~\cite{khashabi2021gooaq} (e.g. \textit{why/how..?}). It can be integrated with a search engine~\cite{metzler2021rethinking}, or a virtual conversation agent, and can also be used to generate explanations as a complement to short-phrase answers for open-domain questions~\cite{kwiatkowski2019natural, yang2018HotpotQA}, or to answer open-ended questions like those from Reddit forum “Explain Like I’m Five”~\cite{fan2019eli5}.


LFQA is quite a challenging task. It often involves searching a large external knowledge source that contains millions of documents for relevant information. Then it \textit{generates} a paragraph-length answer from those retrieved sources. While the great success in retrieval technique~\cite{guu2020realm, karpukhin2020dense, lee2019latent} can be carried over to the LFQA setting, more challenges lie in the generation. First, multiple documents that contain hundreds of tokens need to be considered for generation, raising difficulties in the direct use of current pre-trained language models. Second, as different documents may contain redundant, complementary, or contradictory information, how to synthesis the information and generate a faithful answer that has less hallucinated content is even more challenging.


\begin{figure}[!t]
 \centering
 \vspace{-5pt}
 \includegraphics[width=0.85\linewidth]{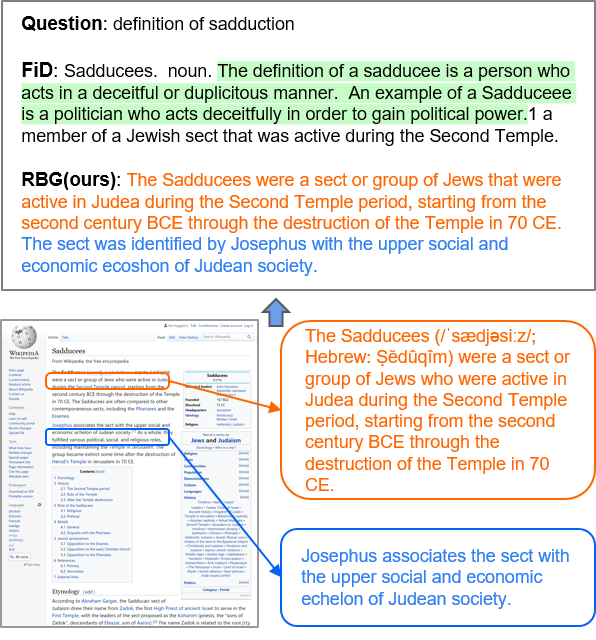}
 \vspace{-10pt}
  \caption{An example from MS MARCO~\cite{nguyen2016ms} dataset. We \hlgreen{highlight} the unfaithful snippets from other model. Our model(\textbf{RBG}) generate more factually accurate answer. }
  \label{fig:example}
  \vspace{-10pt}
\end{figure}

While recent work on LFQA~\cite{krishna2021hurdles} focuses primarily on the first challenge, and has produced fluent and somewhat relevant content, the latter faithfulness challenge has not been explored. However, the faithfulness issue is quite important for LFQA. As the example in Fig.~\ref{fig:example} shown, a fluent and relevant but unfaithful answer (highlight in \hlgreen{green}) will mislead users.


In this paper, we propose a novel end-to-end framework named RBG (\textbf{R}ead \textbf{B}efore \textbf{G}enerate) for LFQA to address the aforementioned challenges. The key idea for enhancing answer faithfulness is to augment the generation process with predicted salient information which can be viewed as an emphasis on answer-related facts. Specifically, we combine a Seq2Seq language model-based generator with a machine reading comprehension (\textit{reader}) module. The \textit{reader} produces an evidence probability score for each sentence, which will be integrated with the generator for final distribution prediction. We perform evidence fusion in a similar way as FiD~\cite{izacard2021leveraging} to equip the pre-trained language model with multiple input documents for generation. To further enhance the factual grounding ability of RBG, we propose an additional pre-training task to encourage the model to rely more on retrieved documents to generate factual statements. The details are explained in Section \ref{sec:methodology}.

We conduct thorough experiments on our method and several baselines on  ELI5~\cite{fan2019eli5}, the only publicly available large-scale LFQA dataset, and also on MS MARCO~\cite{nguyen2016ms} passage ranking data, which can also be transformed into an answer generation task. The proposed method tops the public leaderboard of the KILT~\cite{petroni2021kilt} Benchmark on the ELI5 dataset. It also outperforms the baselines, including non-retrieval and retrieval-based methods, such as DPR-BART~\cite{petronicontext}, RAG~\cite{izacard2021leveraging} and FiD~\cite{izacard2021leveraging}, with improvement on the automatic evaluation results on the MS MARCO dataset. Human evaluation results further validate that our proposed framework can improve the generation quality in terms of relevance and factual correctness. 

Our contributions are summarized as follows:
\begin{itemize}
\setlength{\itemsep}{0pt}
\setlength{\parsep}{0pt}
\setlength{\parskip}{0pt}
\item To the best of our knowledge, we are the first trying to tackle the faithfulness challenge in LFQA.
\item We propose a new and effective framework for open-domain LFQA to generate answers with the guidance of a sentence evidence score from a machine reading module, as well as an additional factual grounding-oriented pre-training task.
\item We show the effectiveness of our method by both automatic evaluation and human evaluation on two large-scale datasets, and we also demonstrate by human evaluation that our method improves the factual correctness of generated answers while still keeping high informativeness.

\end{itemize}
\vspace{-2pt}

\begin{figure*}[!ht]
 \centering
 \includegraphics[width=0.85\linewidth]{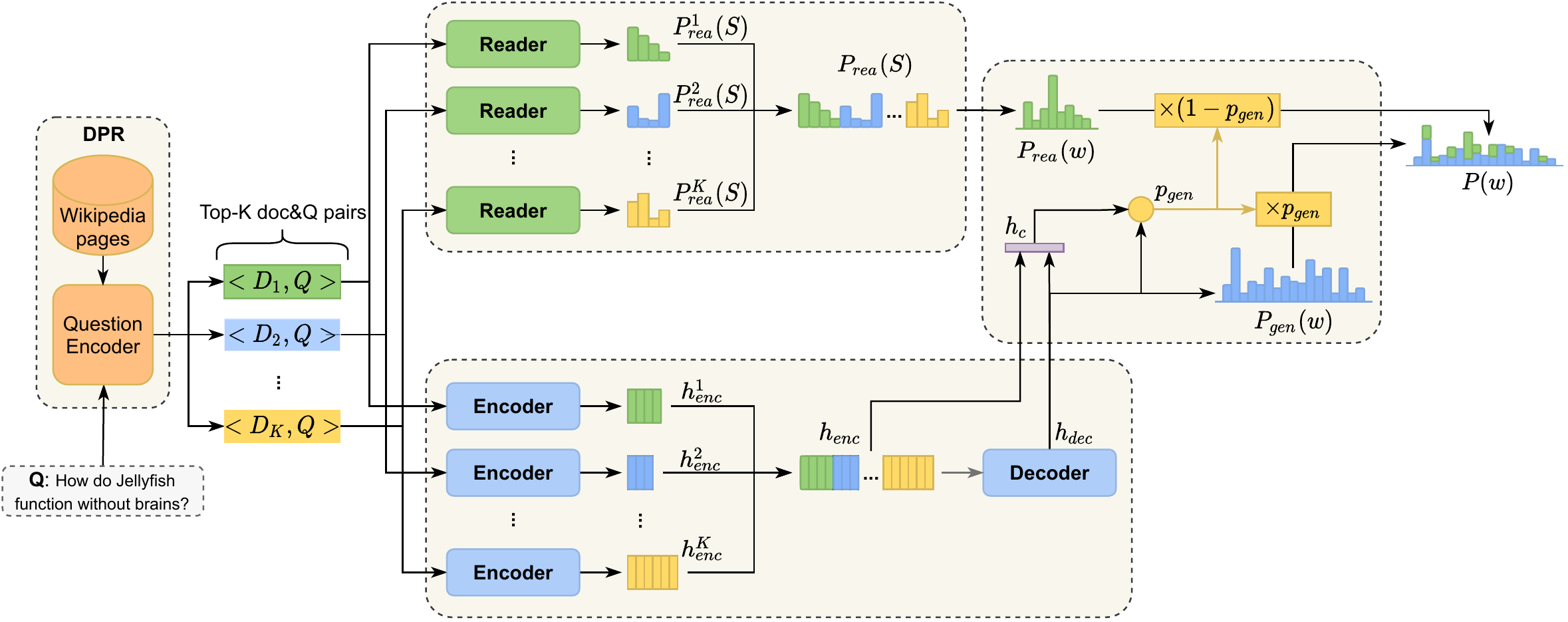}
  \caption{Overview architecture of our RBG framework. RBG comprises a supporting document retriever, a document reader and a generator.}
  \label{fig:overrall model}
\end{figure*}
 \vspace{-10pt}

\section{A state-of-the-art LFQA system}
\label{sec:methodology}
To generate in-depth, long-form answers for a given general domain question, we first use a retriever to search for relevant information from a large external knowledge source. Then our reader and the generation module take the multiple retrieved documents together with the question as input to generate the answer. Specifically, the reader module adopts a machine reading comprehension (MRC) model to produce an evidence score for each sentence in each document, while the generator, which adopts a large pre-trained Seq2Seq language model, fuses the sentence evidence score into its generation process. Our framework is shown in Figure~\ref{fig:overrall model}.

\subsection{Supporting document retriever}

We use DPR~\cite{karpukhin2020dense} to retrieve the supporting documents following the typical methods in the state-of-the-art framework for open-domain QA~\cite{izacard2021leveraging,NEURIPS2020_6b493230}. The passage and question are represented as 768-dimensional dense vector representations, computed via the BERT-based bi-encoder networks of DPR. The retriever will rank the documents according to their relevance, calculated as
\setlength{\abovedisplayskip}{3pt}
\setlength{\belowdisplayskip}{3pt}
\begin{equation}
Score_{re}(Q, D_i) = \text{BERT}_q(Q)^T \text{BERT}_d(D_i)
\label{eqa:retriever_score}
\end{equation}

Retrieval is performed using approximate nearest neighbors with the FAISS\footnote{\url{github.com/facebookresearch/faiss}} library. We denote $D=\{D_1, D_2,..., D_k\}$ as the top-$K$ retrieved documents for question $Q$. 


\subsection{Document reader}
Since there are no golden retrievals for long-form answers, the retrieved documents may contain complementary, contradictory, or redundant information related to the answer. Thus, we propose to use a reader module to explicitly predict the sentence-level evidence probability in each document.  

\paragraph{Evidence span prediction}
We use a machine reading comprehension (MRC) model to predict the evidence span in each document, as these models approach or even outperform human-level performance on many datasets~\cite{joshi2020spanbert}. The MRC model takes the concatenation of the retrieved document $D_i$ and question $Q$ as input, and outputs the prediction of the start and end position of the potential evidence spans in $D_i$. Specifically, it outputs two probability distributions over the tokens in $D_i$: $P^s_i(w_s)$ and $P^e_i(w_s)$, where $P^s_i(w_s)$ / $P^e_i(w_s)$ is the probability that the token $w_s$ is the start/end of the evidence span in $D_i$. 


\vspace{3pt}

\paragraph{Sentence evidence probability}
Originally, the MRC model was designed to give accurate, short-phrase span prediction~\cite{rajpurkar2016squad}, but we argue that a sentence-level evidence probability will be better in our scenario. The supporting sentences can provide the minimum required context information for each answer span, which is quite important, especially in multi-document generation~\cite{xu2020coarse}. We define our sentence-level evidence probability score for the $i$-th document $P^i_{rea}(S)$ as the summation over all token-level evidence probabilities in that sentence, and it is calculated via
\begin{align}
    P^i_{rea}(S) &= \frac{1}{2}\sum\nolimits_{w_s\in S}(P^s_i(w_s) + P^e_i(w_s)) \\
    P_{rea}(S) &= \text{Norm}(P^1_{rea}(S); ...P^i_{rea}(S); ...P^K_{rea}(S)) 
\end{align}
We concatenate $P^i_{rea}$ and normalize the distribution as $P_{rea}(S)$, where $P_{rea}(S)$ denotes the final sentence-level evidence probability in all the $K$ documents regarding the question. 

\paragraph{Multi-task MRC}
As there are no golden answer spans for LFQA data, we need a MRC model that has enough generalization ability for open domain questions as a starting point. We choose SpanBERT~\cite{joshi2020spanbert}, and further fine-tune it in a multi-task way on six large-scale MRC datasets from the MRQA shared task ~\cite{fisch-etal-2019-mrqa} following work by~\citet{su-etal-2019-generalizing}: SQuAD~\cite{rajpurkar2016squad}, NewsQA~\cite{trischler2017newsqa}, TriviaQA~\cite{Joshi_2017}, SearchQA~\cite{dunn2017searchqa}, HotpotQA~\cite{yang2018HotpotQA}, and NatualQuestions~\cite{kwiatkowski2019natural}. The multi-task fine-tuned MRC model $R$ will be further jointly trained with the generator, using the golden answer in a distantly supervised way.

\vspace{-5pt}
\subsection{Generator}

\paragraph{FiD-BART} We choose BART as our generation backbone because of its outstanding performance on many generation tasks, especially on long-form abstractive summarization task~\cite{lewis2020bart}. We propose FiD-BART, following the \textit{Fusion-in-Decoder} idea from~\citet{izacard2021leveraging}, to empower BART to deal with multiple, long-document inputs. FiD-BART processes each document independently in the encoder, while performing the cross-attention in the decoder jointly. 

The encoder encodes the concatenation of each supporting document $D_i$ and the question $Q$. More precisely, we append the special tokens \textit{question:} before $Q$, \textit{title:} and \textit{context:} before the title and text of each document $D_i$. We denote the encoded final representation of the encoder as $h_{enc}$,  which is the concatenation of the $K$ encoder outputs $h^i_{enc}$ ($h^i_{enc} \in R^{d \times l_i}$) for the $i$th document:
\begin{align}
    h^i_{enc} &= \text{Encoder}(Q;D_i) \\
    h_{enc} &= (h^1_{enc},..., h^i_{enc},..., h^K_{enc}) 
\end{align}

The partial structure of the decoder can be illustrated by Eq.(6)--(8), where $h_l$ is the representation for the $l$-th decoder layer. We denote $h_{dec}$ as the last layer decoder outputs: 
\vspace{-3pt}
\begin{align}
    h_l^a &= \text{SelfAttention}(h_l, h_l, h_l) \\
    h_l^b &= \text{LayerNorm}(h_l + h_l^a) \\
\label{cross-attention}    h_l^c &= \text{CrossAttention}(h_l^b, h_{enc}, h_{enc})
\end{align}

\vspace{-3pt}
As we can see, FiD-BART can scale to a large number of input documents within a linear computation time.
\vspace{-7pt}
\subsection{Reader-before-generator}
\vspace{-5pt}
To incorporate the evidence probability into generation, we apply the pointer-generator model (depicted in Figure~\ref{fig:overrall model}). The attention distribution $\mathcal{A}$ and context vector $h_c$, and the generation probability $p_{gen}$ $\in$ [0,1] are calculated as follows:
\begin{align}
    \mathcal{A} &= softmax(h_{dec} h^T_{enc}) \\
    h_{c} &= \mathcal{A}  h_{enc} \\
    p_{gen} &= sigmod(W_{c}h_{c} + W_{g}h_{dec}) 
\end{align}
where $W_c$ and $W_g$ are learnable parameters. $p_{gen}$ is used as a soft switch to choose between generating a word from the generator by sampling from the vocab, or copying a word from the input sequence by sampling according to the evidence distribution $P_{rea}(w)$:
\begin{align}
&P_{gen}(w) = lm_{head}(h_{dec}) \\
&P_{rea}(w) = \sum\nolimits_{s:w_s=w\text{,}w_s \in S}{P_{rea}(S)} \\
&P(w) = p_{gen}P_{gen}(w) + (1-p_{gen})P_{rea}(w)
\end{align}

\vspace{-12pt}
\subsection{Pre-training}
\vspace{-5pt}
To further improve the ability to ground on retrieved documents, we propose a pre-training task: retrieval-augmented recovery~(RAR). Instead of recovering the corrupted text through the internal knowledge memorized in model parameters~\cite{raffel2020exploring,lewis2020bart}, RAR encourages the model to rely more on external retrieved documents to generate factual statements. Specifically, given an original text $S$, we retrieve the top-$k$ documents ${D_1,D_2,...,D_N}$ from the knowledge corpus using BM25~(discarding $S$ itself), and we replace 30\% of the words in $S$ with \mask to form a pseudo query $Q$. The pre-training task asks our RBG model to recover $S$ with the input of the pseudo query $Q$ and $k$ retrieved documents, which can be formulated as
\vspace{-5pt}
\begin{equation}
    S=RBG(Q;D_1,D_2,...,D_k)
    \vspace{-5pt}
\end{equation}
\noindent To involve more factual information during the text corruption and recovery process, we sample 1 million sentences of $S$ corresponding to at least one knowledge base triplet from Wikipedia with the text-triple alignment of TREX~\cite{TREX}.  

\vspace{-10pt}
\section{Experiment Setups}
\vspace{-5pt}

\subsection{Datasets}
We conduct experiments on the two following datasets, both of which concentrate on long form generative QA.
\vspace{-5pt}
\paragraph{ELI5}~\cite{fan2019eli5} is the only publicly available large-scale LFQA dataset. It is a collection of
question-answer pairs extracted from the Reddit forum "Explain Like
I’m Five"(ELI5). We use the KILT~\cite{petroni2021kilt}  version of the dataset from its Github repository\footnote{\url{github.com/facebookresearch/KILT}}, which
has 272,634 training examples and 1,507 development examples. The average length of the answers is 130 words.
\vspace{-5pt}
\paragraph{MS MARCO}~\cite{nguyen2016ms} is a dataset of crowdsourced responses to Bing
queries. We use the question-answer pairs of the MS MARCO passage ranking track for training and evaluation, as they are more abstract and reliant on multi-document information than those of the NLG track. The training example size is about 500,000 and the evaluation example size is 6980. 

\vspace{-5pt}
\paragraph{Knowledge source} The external knowledge source of the retriever is the Wikipedia paragraphs, which are provided in the KILT benchmark as a unified knowledge source for knowledge-intensive tasks, including open-domain LFQA ~\cite{petroni2021kilt}. It is based on the 2019/08/01 Wikipedia snapshot, and contains 5.9M articles.

\vspace{-5pt}
\subsection{Baselines}
\label{sec:baselines}
\paragraph{BART and T5} We fine-tune BART~\cite{lewis2020bart} and T5~\cite{raffel2020exploring} using QA pairs without explicit retrieval, and include them as our baselines which rely only on parameterized internal knowledge~\cite{roberts2020much} to generate answers.
\vspace{-5pt}
\paragraph{DPR-BART} is our retrieval-based LFQA baseline. We follow ~\citet{petronicontext} to retrieve and prepend the top-3 passages from DPR for each input sample, and use context-enhanced training data to fine-tune a BART model.
\vspace{-5pt}
\paragraph{RAG} ~\cite{NEURIPS2020_6b493230} is an end-to-end retrieval-augmented generation model which back-propagates to the retriever’s input encoder. We experiment with fine-tuning RAG on LFQA tasks, establishing a strong baseline on all of them.  At every generation step we retrieve the top-5 passages and use them as supporting documents.
\vspace{-5pt}
\paragraph{FiD} ~\cite{izacard2021leveraging} encodes each passage independently and combines all outputs from the encoder before passing them to the decoder. FiD has achieved superior performance on a number of open-domain QA tasks~\cite{izacard2021leveraging}. We implement FiD-BART, using BART as the generation backbone, as our strongest baseline.

\section{Experiment Results}
\vspace{-3pt}
\subsection{Automatic Evaluation}
\vspace{-3pt}

We use the metrics unigram F1 score and ROUGE-L~\cite{lin2004rouge} in previous work on LFQA~\cite{petroni2021kilt, krishna2021hurdles} to evaluate and compare the generation quality of our method. 

\vspace{-5pt}
\paragraph{Overall Comparison} Table~\ref{results} shows the performance of various methods on the two datasets. As shown, our RBG method outperforms all baselines models with regard to both evaluation metrics on both datasets. The RBG method also outperforms the previous state-of-the-art method \textit{c-REALM+RT} on the KILT-ELI5 leaderboard\footnote{\url{ https://evalai.cloudcv.org/web/challenges/challenge-page/689/leaderboard/1908}} ~\cite{krishna2021hurdles}, as shown in Table~\ref{leaderboard}.

\begin{table}[!ht]
\resizebox{0.47\textwidth}{!}
{
\begin{tabular}{c|cc|cc}
\hline
Models  & \multicolumn{2}{c}{Eli5} &\multicolumn{2}{c}{MS MARCO}\\  
            & ROUGE-L      & F1         & ROUGE-L       & F1 \\ \hline
T5(base)    & 21.02        & 18.36       & 21.19 & 20.03 \\
BART(large) & 22.69        & 22.19      & 23.26 & 25.6  \\
DPR+BART    & 17.41        & 17.88      & 23.01 & 25.13     \\
RAG         & 16.11        & 17.24      &   -    & -  \\
FiD  &    25.70      &    28.55    &   24.64    & 27.08   \\
RBG(ours)   & \textbf{26.46}        & \textbf{29.04}      & \textbf{24.72} & \textbf{27.52} \\ \hline
\end{tabular}
}
\vspace{-5pt}
\caption{Performance comparison between our RBG method and the baselines on the KILT-ELI5~\cite{petroni2021kilt} and MS MARCO~\cite{nguyen2016ms} evaluation sets.}
\label{results}
\vspace{-18pt}
\end{table}

\begin{table}[!ht]
\resizebox{0.47\textwidth}{!}
{

\begin{tabular}{c|ccccc}
\hline
Model           & \multicolumn{2}{c}{Retrieval} & \multicolumn{2}{c}{Generation} &      \\
                & PRr.       & R@5        & F1           & R-L          & KRL  \\ \hline
RBG(ours)       & 10.83         & 27.25         & \textbf{24.53 }         & \textbf{27.13}         & \textbf{2.62} \\
DPR\_kilt\_wiki &        14.83       &   27.69            &      16.45          &       15.91        &    2.46  \\
c-REALM$^1$  &        10.67       &      24.56         &        23.19        &      22.88         &   2.36   \\
DPR+BART        &    10.67           &       26.92        &       17.41         &        17.88       &  1.90    \\
RAG             &        11.00       &      22.92         &      14.05          &      14.51         &   1.69   \\
BART-large      &          0.00     &         0.00     &     20.55            &    19.23            &     0.00  \\ 
T5-base         &      0.00         &       0.00        &       19.08         &      16.10         &   0.00   \\
\hline
\end{tabular}

}
\caption{Results on the ELI5 test set on the KILT leaderboard. Our RBG tops the leaderboard in terms of (1) retrieval performance, using R-precision(RPr.) and Recall@5(R@5), and (2) generation quality, using F1 and ROUGE-L(R-L). These scores are combined to produce the overall metric KILT R-L(KRL)~\cite{petroni2021kilt}. c-REALM$^1$ is from~\cite{krishna2021hurdles}}
\label{leaderboard}
\vspace{-10pt}
\end{table}

\paragraph{Fine-grained Comparison} Intuitively, the quality of retrieved documents will affect the generation quality, thus we provide a fine-grained performance comparison. We split MS-MARCO evaluation set into different subset based on the quality of the retrieved documents\footnote{\label{foot:retrieve} We consider two metrics to measure the retrieval quality for a certain question: (1)~\textbf{Top-1 document retrieval score} which is the matching score output by the retriever (Equation.~\ref{eqa:retriever_score}) for the top-1 document to measure the corresponding semantic relevance to the given question, and (2)~\textbf{N-gram overlap}, which is the N-gram overlap between the golden answer and the top-k retrieved documents.}, and compare the ROUGE-L score between FiD and RBG under each subset.

As we can see from Table~\ref{tab:FidvsRBG}, even though RBG beats FiD by 0.1 Rouge-L score on the whole MS-MARCO evaluation set, the performance gap continue increasing as the retrieval quality of the evaluation subset increased. This indicates that RBG is especially effective when high-quality retrieval documents is provided, which matches with our intuition. 

\begin{table}[!h]
\resizebox{0.45\textwidth}{!}
{
\centering
\begin{tabular}{lc|cccc}
\hline
\multicolumn{2}{l|}{>ngram overlap} & 0     & 0.4        & 0.6       & 0.8    \\ \hline
\multicolumn{2}{l|}{\# of documents} & 6980  & 3493      & 1470      & 489    \\ \hline
\multirow{2}{*}{ROUGE-L}    & FiD    & 24.64 & 28.04  & 33.62  & 45.25 \\
                            & RBG    & 24.72 & 28.59  & 34.38   & 46.29  \\ \hline
\end{tabular}
}
\end{table}
\vspace{-20pt}
\begin{table}[!h]
\resizebox{0.45\textwidth}{!}
{
\begin{tabular}{lc|cccc}
\hline
\multicolumn{2}{l|}{\textgreater{}retrieval score} & 0.0   & 75    & 80    & 85    \\ \hline
\multicolumn{2}{l|}{\# of documents}               & 6980  & 5811  & 3188  & 1001  \\ \hline
\multirow{2}{*}{ROUGE-L}           & FiD           & 24.64 & 24.7  & 25.63 & 26.81 \\
                                   & RBG           & 24.72 & 25.46 & 26.53 & 27.96 \\ \hline
\end{tabular}
}
\vspace{-5pt}
\caption{Fine-grained comparison between FiD and RBG on different subset of MS-MARCO evaluation data.}
\label{tab:FidvsRBG}
\vspace{-10pt}
\end{table}


\vspace{-10pt}
\subsection{Human evaluation} 
We further evaluate our model using human annotators, who we ask to quantify three aspects of the generated answer, (1)~\textbf{fluency}, which measures whether the answer is coherent and less repetitive; (2)~\textbf{relevance}, which measures the amount of information relevant to answering the question, and (3)~\textbf{factual correctness}~(also briefly called correctness), which measures the correctness and faithfulness of all facts involved in the generated answer.

We select FiD, which is the strongest baseline in terms of automatic metrics, for comparison. We sample evaluation questions from the MS MARCO dev set, which are better supported by Wikipedia knowledge than ELI5. Table~\ref{tab:human eval abs} shows the absolute evaluation results of human annotation. To reduce the impact of scale selection inconsistency of different annotators, we also show the relative evaluation results in Table~\ref{tab:human eval rel}. We can see that both types of results indicate that RBG outperforms FiD in terms of all three aspects. RBG has more advantages over FiD on the metric of factual correctness, possibly benefited by the introduction of the reader module and additional pre-training. More details of the human evaluation setup and statistical analysis can be found in Appendix~\ref{appendix:human_eval}.

\begin{table}[]
\resizebox{0.45\textwidth}{!}
{
\begin{tabular}{c|ccc}
\hline
Model    & Fluency      & Relevance      &  Correctness   \\ \hline
FiD  &    2.62      &    2.34    &    2.07    \\
RBG(ours)   & \textbf{2.70}        & \textbf{2.50}     & \textbf{2.41} \\ \hline
\end{tabular}
}
\vspace{-4pt}
\caption{Absolute human evaluation results for RBG vs. FiD on MS MARCO. The table shows the mean value across all annotators and examples for each metric.}
\label{tab:human eval abs}
\vspace{-12pt}
\end{table}

\begin{table}[]
\resizebox{0.45\textwidth}{!}
{
\begin{tabular}{c|ccc}
\hline
Aspect    & Prefer FiD      & Prefer RBG      &  Tie  \\ \hline
Fluency  &    12\%   & \textbf{26\%}   &    62\% \\
Relevance  &  18\%     & \textbf{48\%} & 34\% \\
Correctness &  4\%     & \textbf{62\%} & 34\% \\ \hline
\end{tabular}
}
\vspace{-5pt}
\caption{Relative human evaluation results for RBG vs. FiD on MS MARCO. The percentages represent the ratio of one model being voted as preferred by multiple annotators on a metric.}
\label{tab:human eval rel}
\vspace{-12pt}
\end{table}

\vspace{-5pt}
\subsection{Ablation}
\vspace{-3pt}
To further investigate the contribution and effect of each module in the proposed system, we conducted a systematic ablations on the MS-MARCRO evaluation dataset. 

\begin{table}[!th]
\centering
\resizebox{0.45\textwidth}{!}
{
\begin{tabular}{c|c|cc}
\hline
No. & models     & \multicolumn{2}{c}{MS MARCO} \\ 
  &   & ROUGE-L    & \multicolumn{1}{l}{F1} \\ \hline
0 & RBG(ours)    & 24.72 & 27.52                  \\ 
1 & w/o reader &  24.66 & 27.30 \\
2 & w/o pre-training  & 24.65  & 27.38   \\
3 & w/o reader + pre-training & 24.64  & 27.08 \\
4 & w/ reader frozen &  24.51 & 25.85 \\
5 & w/ random retrieval  & 22.84  & 25.23  \\
\hline
\end{tabular}
}
\vspace{-5pt}
\caption{Ablation results on the MS MARCO evaluation set. A more fine-grained results comparison is shown with analysis in Section ~\ref{sec:further}.}
\label{ablation}
\vspace{-15pt}
\end{table}

\vspace{-0.5pt}
\paragraph{\textit{w/o} reader/pre-training:} We respectively remove the reader module (\textbf{w/o reader}), the pre-training (\textbf{w/o pre-training}), and both together (\textbf{w/o reader + pre-training}) from our model , to test the contribution of each part. As we can see from Table~\ref{ablation}, without the reader to predict the evidence probability, the generation performance decreases in both metrics, and the performance continues to drop without the pre-training. 

\vspace{-5pt}
\paragraph{\textit{w/} reader frozen:}  We freeze the reader to investigate the benefit of distantly supervised end-to-end training of the reader module. As we can see from Table~\ref{ablation}, the results on both metrics drop, especially the F1 score, which proves the effectiveness of the end-to-end training.
\vspace{-5pt}
\paragraph{\textit{w/} random retrieval:} To investigate whether and how much the generation process is grounded in the retrieved documents, we replace retrieved paragraphs with randomly sampled paragraphs from Wikipedia at \textit{inference} time for comparison. As we can see, the ROUGE-L drops significantly with randomly retrieved documents, and it is also worse than the baseline systems such as BART and DPR-BART (Table~\ref{results}).

\vspace{-5pt}
\section{Further analysis} 
\vspace{-5pt}
\label{sec:further}
We conduct further analysis on the results, considering that LFQA is a complicated but less explored task, which deserves a complete investigation. 
\vspace{-5pt}
\subsection{How does retriever affect the generation quality?}
\vspace{-3pt}
We further investigate the effects of the quality of retrieved documents on the final generation. We split the evaluation sets of the two datasets via different thresholds for the two metrics\textsuperscript{\ref{foot:retrieve}} and calculate the corresponding ROUGE-L score for each subset. As we can see in Table~\ref{retrieval_results}, better-retrieved documents always bring better generation quality, indicating the importance of high-quality supporting documents for the generation process.

We also measure the effects of the number of retrieved documents $K$ on the generation quality and find that the best $K$ from $\{5,10,20,50\}$ is 10. More retrieved documents do not improve generation quality as in open-domain QA.

\begin{table}[!h]
\resizebox{0.45\textwidth}{!}
{
\begin{tabular}{c|cc|cc}
\hline
\multicolumn{1}{l|}{\multirow{2}{*}{\begin{tabular}[c]{@{}l@{}}\textgreater{}retrieval\\ score(top-1)\end{tabular}}} & \multicolumn{2}{c|}{ELI5}                                    & \multicolumn{2}{c}{MS MARCO}                                 \\ \cline{2-5} 
\multicolumn{1}{l|}{}                                                                                                & \multicolumn{1}{l}{\# of data} & \multicolumn{1}{l|}{ROUGE-L} & \multicolumn{1}{l}{\# of data} & \multicolumn{1}{l}{ROUGE-L} \\ \hline
0.0 & 1570  & 26.35  & 6980  & 24.72 \\
75 & 1270 & 26.37 & 5811  & 25.46                      \\
80 & 479  & 26.38 & 3188  & 26.53                      \\
85  & 72  & 26.96   & 1001     & 27.96                      \\
90 & 11    & 27.25   & 161  & 27.61                      \\ \hline
\end{tabular}
}
\end{table}
\vspace{-18pt}
\begin{table}[!h]
\resizebox{0.45\textwidth}{!}
{
\begin{tabular}{c|cc|cc}
\hline
\multicolumn{1}{l|}{\multirow{2}{*}{\begin{tabular}[c]{@{}l@{}}\textgreater{}ngram \\ overlap\end{tabular}}} & \multicolumn{2}{c|}{ELI5}                                    & \multicolumn{2}{c}{MS MARCO}                                 \\ \cline{2-5} 
\multicolumn{1}{l|}{}                                                                                        & \multicolumn{1}{l}{\# of data} & \multicolumn{1}{l|}{ROUGE-L} & \multicolumn{1}{l}{\# of data} & \multicolumn{1}{l}{ROUGE-L} \\ \hline
0.0  & 1570   & 26.35  & 6980  & 24.72 \\
0.4 & 460 & 27.09& 3493 & 28.59  \\
0.5 & 260& 27.31 & 2470 & 30.72 \\
0.6 & 109  & 27.52 & 1470 & 34.38 \\
0.7 & 48 & 27.63 & 845  & 39.64  \\
0.8& 27  & 27.17& 489  & 46.29  \\ \hline
\end{tabular}
}
\vspace{-5pt}
\caption{Fine-grained results of our RBG on ELI5 and MS MARCO. With high-quality retrieval (higher N-gram overlap or retrieval score threshold), the answer quality (ROUGE-L) increases on both datasets.}
\label{retrieval_results}
\end{table}
\vspace{-10pt}

\vspace{-5pt}
\subsection{How does the reader contribute to the generation?}
\vspace{-5pt}

As shown in the ablation study, the reader module improves the overall performance on the MS MARCO evaluation dataset. We further investigate its performance when retrieved documents with different quality levels are provided. 

\begin{figure}[!t]
\vspace{-5pt}
	\begin{center}
		\includegraphics[width=0.46\textwidth]{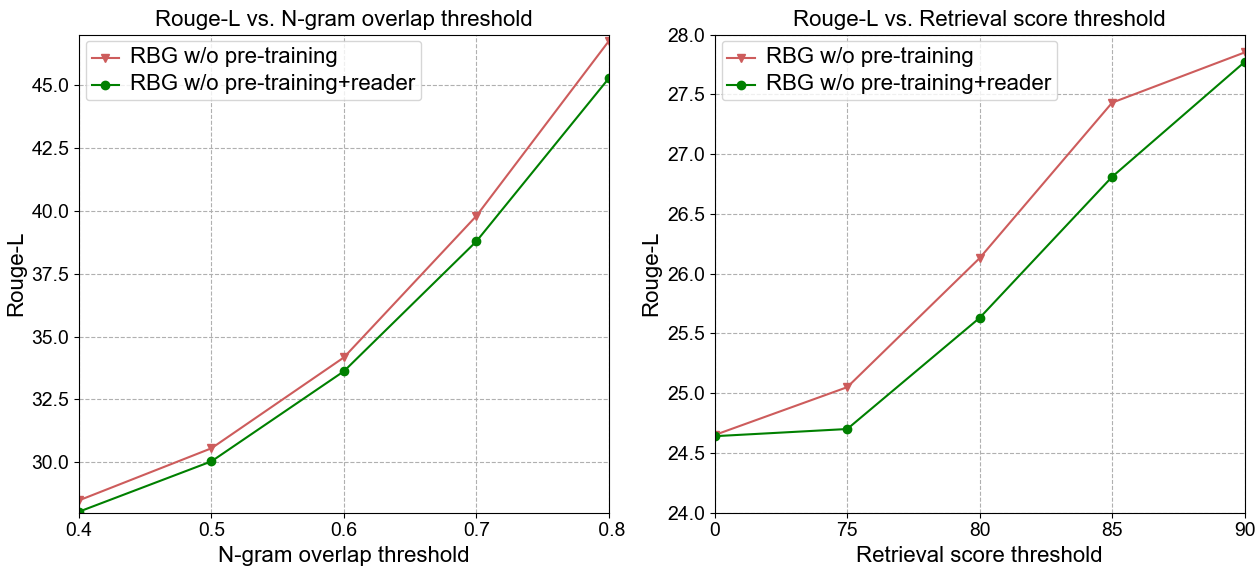}
		\vspace{-5pt}
		\caption{ROUGE-L versus document retrieval performance for reader analysis.}
		\label{Fig:decomp_for_reader}
	\end{center}
	\vspace{-0.2cm}
\end{figure}

We show in Figure~\ref{Fig:decomp_for_reader} the fine-grained comparison results between ablation models  \textbf{No.2}: \textit{RBG w/o pre-training} and \textbf{No.3}: \textit{RBG w/o pre-training + reader}. As we can see, the difference in ROUGE-L between the two models increases as the quality of the retrieved documents improves, indicating the reader's strong capability, especially on high-quality data. This also matches with our intuition. We also conduct a human evaluation for reader analysis, and we show the results in Table~\ref{tab:human eval reader}. 

\begin{table}[]
\resizebox{0.48\textwidth}{!}
{
\begin{tabular}{c|ccc}
\hline
Aspect    & Prefer w/o reader      & Prefer w/ reader      &  Tie \\ \hline
Fluency  &    15\%    & \textbf{35\%}   &    50\% \\
Relevance  &  17\%     & \textbf{57\%} &26\% \\
Correctness &  25\%     & \textbf{45\%} & 30\% \\ \hline
\end{tabular}
}
\vspace{-5pt}
\caption{Human evaluation results for RBG reader analysis on MS MARCO. The model with reader has better generation performance in terms of fluency, relevance and correctness.}
\label{tab:human eval reader}
\vspace{-12pt}
\end{table}

\begin{figure}[t]
\vspace{-5pt}
	\begin{center}
		\includegraphics[width=0.46\textwidth]{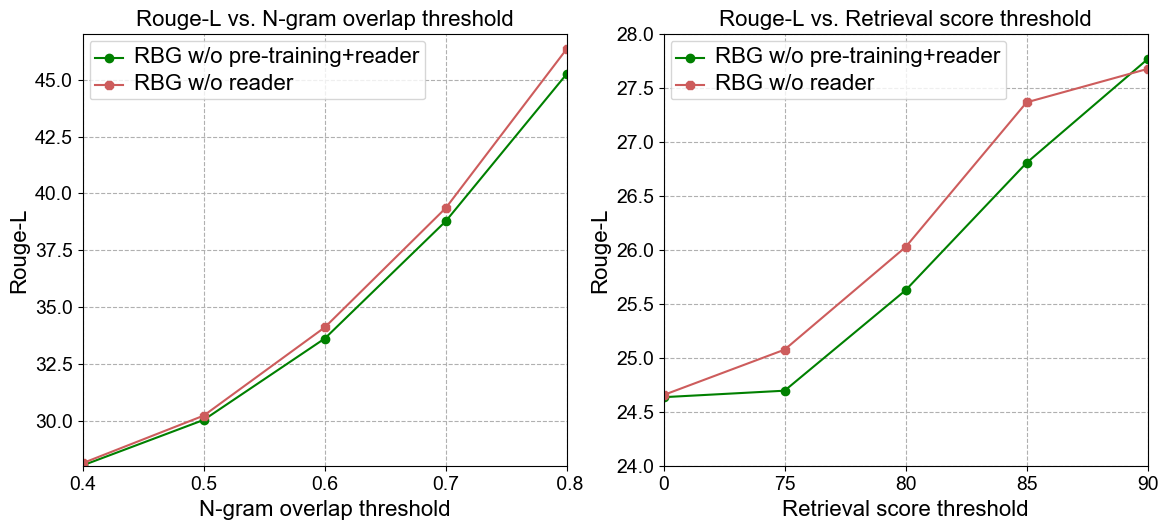}
		\vspace{-5pt}
		\caption{ROUGE-L versus Document retrieval performance for pre-training analysis.}
		\label{Fig:decomp_for_pretrain}
	\end{center}
	\vspace{-10pt}
\end{figure}

\vspace{-5pt}
\subsection{How does pre-training help?}
\vspace{-5pt}

\begin{table}[]
\vspace{-10pt}
\resizebox{0.48\textwidth}{!}
{
\begin{tabular}{c|ccc}
\hline
Aspect    & Prefer w/o pre-training      & Prefer w/ pre-training      &  Tie \\ \hline
Fluency  &    40\%    & \textbf{43\%}   &    17\% \\
Relevance  &  20\%     & \textbf{33\%} & 47\% \\
Correctness &  23\%     & \textbf{47\%} & 30\% \\ \hline
\end{tabular}
}
\vspace{-5pt}
\caption{Human evaluation results for RBG pre-training analysis on MS MARCO. The model with RAR pre-training has better generation performance in terms of relevance and correctness.}
\label{tab:human eval pretrain}
\vspace{-10pt}
\end{table}

We also compare the models' performance in a fine-grained way, to quantify the contribution from our pre-training task. We show in Figure~\ref{Fig:decomp_for_pretrain} the fine-grained comparison results between ablation models \textbf{No.1}: \textit{RBG w/o reader} and \textbf{No.3}: \textit{RBG w/o pre-training + reader}. As we can see, the model with pre-training is better in most situations than that without pre-training. The human evaluation in Table~\ref{tab:human eval pretrain} also indicates the effectiveness of our pre-training task to improve the factual correctness and relevance of the generated answer. We conjecture that the pre-training task of retrieval-augmented recovery can facilitate the downstream LFQA model to combine multiple pieces of evidence from different retrieved documents to generate the final answer.

\vspace{-5pt}
\subsection{Faithfulness analysis}
\vspace{-3pt}

\paragraph{Zero-shot on extractive QA tasks}  Inspired by previous work~\cite{wang2020asking, durmus2020feqa} which leverage a Question Generation(QG) and a QA model to generate question answer pairs, to evaluate the faithfulness of a summary\footnote{
They generate question answer pairs <$q$,$a_{sum}$> from the summary, and compare $a_{sum}$ with the answer $a_{sc}$ from source document for $q$, to evaluate faithfulness.}, we propose to evaluate answer faithfulness via evaluation on two simpler open-domain QA datasets: NaturalQuestions~\cite{kwiatkowski2019natural} and HotpotQA~\cite{yang2018HotpotQA}, which contain single-hop or multi-hop factual questions with golden answers ($\{(q_i, a^s_i)\}_{i=1}^m$) where $a_i^s$ can be extracted from Wikipedia-based documents. We use the trained models (based on MS MARCO) in Table~\ref{results} to do zero-shot long-form answer generation for these two datasets $\{a^l_i = \text{Model}_{\text{ms}}(q_i)\}$, and measure the short-answer recall~(the ratio of golden answer span $a^s$ contained in the generated long answer $a^l$) as an estimation of faithfulness of the generated long-answer:
\begin{equation}
Score(q, a^s, a^l) = \frac{\sum_{i=1}^m\mathbbm{1}[a^s_i \in a^l_i]}{m}
\label{eqa:faithfulness}
\end{equation}
We show the results in Table~\ref{nq_HotpotQA}. As we can see, our system achieves comparable performance with FiD on NQ, and it consistently outperforms other strong baselines on multi-hop dataset hotpotQA, indicating its capability in generating faithful answer especially on complex question that need to synthesis information. We also give concrete examples in Appendix~\ref{appendix:results} that show our model can generate more faithful snippets than FiD apart from automatic metrics. 

\begin{table}[!t]
\centering
\resizebox{0.40\textwidth}{!}
{
\begin{tabular}{c|cc}
\hline
          & NQ Recall   & HotpotQA Recall\\ \hline
T5         & 4.76  & 7.20     \\
BART-large & 10.44 & 9.13   \\
DPR+BART   & 16.37 & 11.57     \\
FiD      & 43.93 & 22.94    \\ \hline
RBG(ours)  & 43.93 & 23.36    \\ \hline
\end{tabular}
}
\vspace{-5pt}
\caption{Faithfulness Analysis of the system generation quality via zero-shot evaluation on NQ~\cite{kwiatkowski2019natural} and HotpotQA~\cite{yang2018HotpotQA}.}
\vspace{-10pt}
\label{nq_HotpotQA}
\vspace{-5pt}
\end{table}

\vspace{-6pt}
\paragraph{Case Study} To have a concrete understanding of the reader's role to address faithfulness, we show two examples in Table~\ref{tab:reader_examples}. While both models use the same \textbf{ctxs}, RBG \textbf{w reader} generates a more \textit{relevant} answer, and \textbf{w/o reader} only gives some correct but unrelated statements (Case 1). The reader also contributes to generating more \textit{faithful} answers, as shown in Case 2. However, we notice that there is one unfaithful statement, which hallucinates the 'second-oldest' as 'oldest'. This might be because of the property of our MRC model: it gives a high evidential score to one sentence, and weakening the importance of other sentences in the same document.

\begin{table}[!ht]
\vspace{-5pt}
\small
\begin{adjustbox}{width={0.49\textwidth},totalheight={\textheight},keepaspectratio}
\begin{tabular}{p{1\columnwidth}}
\hline
\textbf{Question:} what does abc stand for in liquor store?  \\
\hline
\textbf{w/o reader answer:} ABC stores. A liquor store is a retail shop that predominantly sells prepackaged alcoholic beverages intended to be consumed off the store's premises. ...,or other similar terms. \\
\textbf{w/ reader answer}: ABC stands for Alcoholic Beverages Control (ABC Stores) and is a type of liquor store that is owned and operated exclusively by the state government, where liquor stores often sell only spirits or sometimes sell spirits and wine but not beer...\\
\textbf{ctxs:} [\{text: ...\textcolor[RGB]{5, 30, 255}{In seventeen alcoholic beverage control (ABC) states}, ...\},...,\{text:\textcolor[RGB]{5, 30, 255}{operated exclusively by the state government, where liquor stores often sell only spirits or sometimes sell spirits and wine but not beer.}, ...,because purchased liquor must be packaged in sealed bottles\},...,\{...\}]\\
\hline
\textbf{Question:} dow jones industrial average please? \\
\hline
\textbf{w/o reader answer:} The Dow Jones Industrial Average (DJIA) is an index of \hlgreen{30 major U.S. stock indexes}. The DJIA was created in 1896 by Charles Dow, a statistician and editor of The Wall Street Journal. \hlgreen{The index was first published in the mid-1880s, and was last updated on May 26, 1896.} \\
\textbf{w/ reader answer:} The Dow Jones Industrial Average (DJIA) is a stock market index that tracks the performance of the 30 largest U.S. industrial companies. It is owned by S\&P Dow Jones Indices, which is majority owned by Standard \& Poor's Global. The Dow is the \hlgreen{oldest} and most famous of the Dow averages. It was created by Charles Dow in 1896. \\
 \textbf{ctxs:} [\{text:...,was originally published on February 16, 1885. ...\textcolor[RGB]{5, 30, 255}{The industrial average was first calculated on May 26, 1896}..\},...\{text:...It is the \textbf{second-oldest} U.S. market index after the Dow Jones Transportation Average. \textcolor[RGB]{5, 30, 255}{Currently owned by S\&P Dow Jones Indices, which is majority owned by S\&P Global}..\},...,\{...\}]
 \\
 \hline
\end{tabular}
 \end{adjustbox}
\caption{Examples from MS MARCO dataset. We \textcolor[RGB]{5, 30, 255}{highlight} the sentences that have high evidential probability from the reader, and use \hlgreen{green} to mark out the unfaithful snippets. }
\label{tab:reader_examples}
\vspace{-15pt}
\end{table}

\section{Related work}
\vspace{-5pt}
\paragraph{Grounded generation} is the task of leveraging external knowledge sources to enhance the generation. Previous work has either used \textit{structured} external knowledge source ~\cite{liu2018knowledge,young2018augmenting,su2020multi} or \textit{unstructured} data. ~\citet{zhou2018dataset} introduced a document grounded dataset for text conversations, and ~\citet{wu2021controllable} proposed to extract lexical control phrases to do controllable grounded response generation, while ~\citet{zhang2021joint} jointly trained a retriever and generator so that annotated text reference parallel data are not needed. 

\vspace{-5pt}
\paragraph{Open-domain QA} is the task of answering general-domain questions~\cite{chen2017reading}, in which the evidence is usually not given. Models that explicitly exploit an external corpus are referred as \textit{open-book} models~\cite{roberts2020much}. They typically index the corpus and then \textit{retrieve-and-read} to extract the answer span from documents~\cite{chen2017reading, lee2019latent, izacard2021leveraging,NEURIPS2020_6b493230}. Another recently proposed class of methods is \textit{closed-book} QA models~\cite{ye2020studying,roberts2020much}. They fine-tune pre-trained language models such as T5~\cite{raffel2020exploring} or BART~\cite{lewis2020bart} with QA pairs without access to any external knowledge or context.

\vspace{-5pt}
\paragraph{Query driven multi-document summarization} (QFMD) aims to generate a summary according to the query and the provided relevant document(s)~\cite{tombros1998advantages}. ~\citet{baumel2018query} incorporated query relevance into a pre-trained abstractive summarizer, while~\citet{xu2020coarse} and ~\citet{su2020caire} utilized QA models for sentence- or paragraph- level evidence ranking.~\citet{su-etal-2021-improve} tried to improve the relevance of the summary by incorporating an answer relevance score for the source documents into the generation. 

\vspace{-5pt}
\section{Conclusion}
\vspace{-5pt}
We propose a new end-to-end framework RBG that jointly models answer generation and machine reading to tackle the faithfulness issue in LFQA. Experiments on two LFQA datasets, ELI5 and MS MARCO, demonstrate the effectiveness of our method in comparison with strong baselines on automatic and human evaluation metrics. The detailed analysis further proves the competency of our method in generating fluent, relevant, and more faithful answers. We also propose to evaluate the factual correctness of LFQA model by answering questions of extractive QA tasks~(e.g., Natural Questions), which may be helpful to evaluate the faithfulness of LFQA model efficiently.










\normalem
\bibliography{anthology,custom}

\begin{thebibliography}{49}
\expandafter\ifx\csname natexlab\endcsname\relax\def\natexlab#1{#1}\fi

\bibitem[{Baumel et~al.(2018)Baumel, Eyal, and Elhadad}]{baumel2018query}
Tal Baumel, Matan Eyal, and Michael Elhadad. 2018.
\newblock Query focused abstractive summarization: Incorporating query
  relevance, multi-document coverage, and summary length constraints into
  seq2seq models.
\newblock \emph{arXiv preprint arXiv:1801.07704}.

\bibitem[{Chen et~al.(2017)Chen, Fisch, Weston, and Bordes}]{chen2017reading}
Danqi Chen, Adam Fisch, Jason Weston, and Antoine Bordes. 2017.
\newblock Reading wikipedia to answer open-domain questions.
\newblock In \emph{Proceedings of the 55th Annual Meeting of the Association
  for Computational Linguistics (Volume 1: Long Papers)}, pages 1870--1879.

\bibitem[{Dinan et~al.(2018)Dinan, Roller, Shuster, Fan, Auli, and
  Weston}]{dinan2018wizard}
Emily Dinan, Stephen Roller, Kurt Shuster, Angela Fan, Michael Auli, and Jason
  Weston. 2018.
\newblock Wizard of wikipedia: Knowledge-powered conversational agents.
\newblock In \emph{International Conference on Learning Representations}.

\bibitem[{Dunn et~al.(2017)Dunn, Sagun, Higgins, Guney, Cirik, and
  Cho}]{dunn2017searchqa}
Matthew Dunn, Levent Sagun, Mike Higgins, V~Ugur Guney, Volkan Cirik, and
  Kyunghyun Cho. 2017.
\newblock Searchqa: A new q\&a dataset augmented with context from a search
  engine.
\newblock \emph{arXiv preprint arXiv:1704.05179}.

\bibitem[{Durmus et~al.(2020)Durmus, He, and Diab}]{durmus2020feqa}
Esin Durmus, He~He, and Mona Diab. 2020.
\newblock Feqa: A question answering evaluation framework for faithfulness
  assessment in abstractive summarization.
\newblock In \emph{Proceedings of the 58th Annual Meeting of the Association
  for Computational Linguistics}, pages 5055--5070.

\bibitem[{Elsahar et~al.(2018{\natexlab{a}})Elsahar, Vougiouklis, Remaci,
  Gravier, Hare, Laforest, and Simperl}]{TREX}
Hady Elsahar, Pavlos Vougiouklis, Arslen Remaci, Christophe Gravier, Jonathon
  Hare, Frederique Laforest, and Elena Simperl. 2018{\natexlab{a}}.
\newblock T-rex: A large scale alignment of natural language with knowledge
  base triples.
\newblock In \emph{Proceedings of the Eleventh International Conference on
  Language Resources and Evaluation (LREC 2018)}.

\bibitem[{Elsahar et~al.(2018{\natexlab{b}})Elsahar, Vougiouklis, Remaci,
  Gravier, Hare, Laforest, and Simperl}]{elsahar2018t}
Hady Elsahar, Pavlos Vougiouklis, Arslen Remaci, Christophe Gravier, Jonathon
  Hare, Frederique Laforest, and Elena Simperl. 2018{\natexlab{b}}.
\newblock T-rex: A large scale alignment of natural language with knowledge
  base triples.
\newblock In \emph{Proceedings of the Eleventh International Conference on
  Language Resources and Evaluation (LREC 2018)}.

\bibitem[{Fan et~al.(2019)Fan, Jernite, Perez, Grangier, Weston, and
  Auli}]{fan2019eli5}
Angela Fan, Yacine Jernite, Ethan Perez, David Grangier, Jason Weston, and
  Michael Auli. 2019.
\newblock Eli5: Long form question answering.
\newblock In \emph{Proceedings of the 57th Annual Meeting of the Association
  for Computational Linguistics}, pages 3558--3567.

\bibitem[{Fisch et~al.(2019)Fisch, Talmor, Jia, Seo, Choi, and
  Chen}]{fisch-etal-2019-mrqa}
Adam Fisch, Alon Talmor, Robin Jia, Minjoon Seo, Eunsol Choi, and Danqi Chen.
  2019.
\newblock \href {https://doi.org/10.18653/v1/D19-5801} {{MRQA} 2019 shared
  task: Evaluating generalization in reading comprehension}.
\newblock In \emph{Proceedings of the 2nd Workshop on Machine Reading for
  Question Answering}, pages 1--13, Hong Kong, China. Association for
  Computational Linguistics.

\bibitem[{Fleiss(1971)}]{fleiss_kappa}
Joseph~L Fleiss. 1971.
\newblock Measuring nominal scale agreement among many raters.
\newblock \emph{Psychological bulletin}, page 378.

\bibitem[{Guu et~al.(2020)Guu, Lee, Tung, Pasupat, and Chang}]{guu2020realm}
Kelvin Guu, Kenton Lee, Zora Tung, Panupong Pasupat, and Ming-Wei Chang. 2020.
\newblock Realm: Retrieval-augmented language model pre-training.
\newblock \emph{arXiv preprint arXiv:2002.08909}.

\bibitem[{Izacard and Grave(2021)}]{izacard2021leveraging}
Gautier Izacard and {\'E}douard Grave. 2021.
\newblock Leveraging passage retrieval with generative models for open domain
  question answering.
\newblock In \emph{Proceedings of the 16th Conference of the European Chapter
  of the Association for Computational Linguistics: Main Volume}, pages
  874--880.

\bibitem[{Joshi et~al.(2020)Joshi, Chen, Liu, Weld, Zettlemoyer, and
  Levy}]{joshi2020spanbert}
Mandar Joshi, Danqi Chen, Yinhan Liu, Daniel~S Weld, Luke Zettlemoyer, and Omer
  Levy. 2020.
\newblock Spanbert: Improving pre-training by representing and predicting
  spans.
\newblock \emph{Transactions of the Association for Computational Linguistics},
  8:64--77.

\bibitem[{Joshi et~al.(2017)Joshi, Choi, Weld, and Zettlemoyer}]{Joshi_2017}
Mandar Joshi, Eunsol Choi, Daniel Weld, and Luke Zettlemoyer. 2017.
\newblock \href {https://doi.org/10.18653/v1/p17-1147} {Triviaqa: A large scale
  distantly supervised challenge dataset for reading comprehension}.
\newblock \emph{Proceedings of the 55th Annual Meeting of the Association for
  Computational Linguistics (Volume 1: Long Papers)}.

\bibitem[{Karpukhin et~al.(2020)Karpukhin, Oguz, Min, Lewis, Wu, Edunov, Chen,
  and Yih}]{karpukhin2020dense}
Vladimir Karpukhin, Barlas Oguz, Sewon Min, Patrick Lewis, Ledell Wu, Sergey
  Edunov, Danqi Chen, and Wen-tau Yih. 2020.
\newblock Dense passage retrieval for open-domain question answering.
\newblock In \emph{Proceedings of the 2020 Conference on Empirical Methods in
  Natural Language Processing (EMNLP)}, pages 6769--6781.

\bibitem[{Khashabi et~al.(2021)Khashabi, Ng, Khot, Sabharwal, Hajishirzi, and
  Callison-Burch}]{khashabi2021gooaq}
Daniel Khashabi, Amos Ng, Tushar Khot, Ashish Sabharwal, Hannaneh Hajishirzi,
  and Chris Callison-Burch. 2021.
\newblock Gooaq: Open question answering with diverse answer types.
\newblock \emph{arXiv preprint arXiv:2104.08727}.

\bibitem[{Kingma and Ba(2014)}]{kingma2014adam}
Diederik~P Kingma and Jimmy Ba. 2014.
\newblock Adam: A method for stochastic optimization.
\newblock \emph{arXiv preprint arXiv:1412.6980}.

\bibitem[{Krishna et~al.(2021)Krishna, Roy, and Iyyer}]{krishna2021hurdles}
Kalpesh Krishna, Aurko Roy, and Mohit Iyyer. 2021.
\newblock Hurdles to progress in long-form question answering.
\newblock In \emph{Proceedings of the 2021 Conference of the North American
  Chapter of the Association for Computational Linguistics: Human Language
  Technologies}, pages 4940--4957.

\bibitem[{Kwiatkowski et~al.(2019)Kwiatkowski, Palomaki, Redfield, Collins,
  Parikh, Alberti, Epstein, Polosukhin, Devlin, Lee
  et~al.}]{kwiatkowski2019natural}
Tom Kwiatkowski, Jennimaria Palomaki, Olivia Redfield, Michael Collins, Ankur
  Parikh, Chris Alberti, Danielle Epstein, Illia Polosukhin, Jacob Devlin,
  Kenton Lee, et~al. 2019.
\newblock Natural questions: A benchmark for question answering research.
\newblock \emph{Transactions of the Association for Computational Linguistics},
  7:452--466.

\bibitem[{Landis and Koch(1977)}]{landis1977measurement}
J~Richard Landis and Gary~G Koch. 1977.
\newblock The measurement of observer agreement for categorical data.
\newblock \emph{biometrics}.

\bibitem[{Lee et~al.(2019)Lee, Chang, and Toutanova}]{lee2019latent}
Kenton Lee, Ming-Wei Chang, and Kristina Toutanova. 2019.
\newblock Latent retrieval for weakly supervised open domain question
  answering.
\newblock In \emph{Proceedings of the 57th Annual Meeting of the Association
  for Computational Linguistics}, pages 6086--6096.

\bibitem[{Levy et~al.(2017)Levy, Seo, Choi, and Zettlemoyer}]{levy2017zero}
Omer Levy, Minjoon Seo, Eunsol Choi, and Luke Zettlemoyer. 2017.
\newblock Zero-shot relation extraction via reading comprehension.
\newblock In \emph{Proceedings of the 21st Conference on Computational Natural
  Language Learning (CoNLL 2017)}, pages 333--342.

\bibitem[{Lewis et~al.(2020{\natexlab{a}})Lewis, Liu, Goyal, Ghazvininejad,
  Mohamed, Levy, Stoyanov, and Zettlemoyer}]{lewis2020bart}
Mike Lewis, Yinhan Liu, Naman Goyal, Marjan Ghazvininejad, Abdelrahman Mohamed,
  Omer Levy, Veselin Stoyanov, and Luke Zettlemoyer. 2020{\natexlab{a}}.
\newblock Bart: Denoising sequence-to-sequence pre-training for natural
  language generation, translation, and comprehension.
\newblock In \emph{Proceedings of the 58th Annual Meeting of the Association
  for Computational Linguistics}, pages 7871--7880.

\bibitem[{Lewis et~al.(2020{\natexlab{b}})Lewis, Perez, Piktus, Petroni,
  Karpukhin, Goyal, K\"{u}ttler, Lewis, Yih, Rockt\"{a}schel, Riedel, and
  Kiela}]{NEURIPS2020_6b493230}
Patrick Lewis, Ethan Perez, Aleksandra Piktus, Fabio Petroni, Vladimir
  Karpukhin, Naman Goyal, Heinrich K\"{u}ttler, Mike Lewis, Wen-tau Yih, Tim
  Rockt\"{a}schel, Sebastian Riedel, and Douwe Kiela. 2020{\natexlab{b}}.
\newblock \href
  {https://proceedings.neurips.cc/paper/2020/file/6b493230205f780e1bc26945df7481e5-Paper.pdf}
  {Retrieval-augmented generation for knowledge-intensive nlp tasks}.
\newblock In \emph{Advances in Neural Information Processing Systems},
  volume~33, pages 9459--9474. Curran Associates, Inc.

\bibitem[{LIN(2004)}]{lin2004rouge}
C-Y LIN. 2004.
\newblock Rouge: A package for automatic evaluation of summaries.
\newblock In \emph{Proc. of Workshop on Text Summarization Branches Out, Post
  Conference Workshop of ACL 2004}.

\bibitem[{Liu et~al.(2018)Liu, Chen, Ren, Feng, Liu, and
  Yin}]{liu2018knowledge}
Shuman Liu, Hongshen Chen, Zhaochun Ren, Yang Feng, Qun Liu, and Dawei Yin.
  2018.
\newblock Knowledge diffusion for neural dialogue generation.
\newblock In \emph{Proceedings of the 56th Annual Meeting of the Association
  for Computational Linguistics (Volume 1: Long Papers)}, pages 1489--1498.

\bibitem[{Metzler et~al.(2021)Metzler, Tay, Bahri, and
  Najork}]{metzler2021rethinking}
Donald Metzler, Yi~Tay, Dara Bahri, and Marc Najork. 2021.
\newblock Rethinking search: Making experts out of dilettantes.
\newblock \emph{arXiv preprint arXiv:2105.02274}.

\bibitem[{Nakano et~al.(2021)Nakano, Hilton, Balaji, Wu, Ouyang, Kim, Hesse,
  Jain, Kosaraju, Saunders et~al.}]{nakano2021webgpt}
Reiichiro Nakano, Jacob Hilton, Suchir Balaji, Jeff Wu, Long Ouyang, Christina
  Kim, Christopher Hesse, Shantanu Jain, Vineet Kosaraju, William Saunders,
  et~al. 2021.
\newblock Webgpt: Browser-assisted question-answering with human feedback.
\newblock \emph{arXiv preprint arXiv:2112.09332}.

\bibitem[{Nguyen et~al.(2016)Nguyen, Rosenberg, Song, Gao, Tiwary, Majumder,
  and Deng}]{nguyen2016ms}
Tri Nguyen, Mir Rosenberg, Xia Song, Jianfeng Gao, Saurabh Tiwary, Rangan
  Majumder, and Li~Deng. 2016.
\newblock Ms marco: A human generated machine reading comprehension dataset.
\newblock In \emph{CoCo@ NIPS}.

\bibitem[{Petroni et~al.(2020)Petroni, Lewis, Piktus, Rockt{\"a}schel, Wu,
  Miller, and Riedel}]{petronicontext}
Fabio Petroni, Patrick Lewis, Aleksandra Piktus, Tim Rockt{\"a}schel, Yuxiang
  Wu, Alexander~H Miller, and Sebastian Riedel. 2020.
\newblock How context affects language models’ factual predictions.

\bibitem[{Petroni et~al.(2021)Petroni, Piktus, Fan, Lewis, Yazdani, De~Cao,
  Thorne, Jernite, Karpukhin, Maillard et~al.}]{petroni2021kilt}
Fabio Petroni, Aleksandra Piktus, Angela Fan, Patrick Lewis, Majid Yazdani,
  Nicola De~Cao, James Thorne, Yacine Jernite, Vladimir Karpukhin, Jean
  Maillard, et~al. 2021.
\newblock Kilt: a benchmark for knowledge intensive language tasks.
\newblock In \emph{Proceedings of the 2021 Conference of the North American
  Chapter of the Association for Computational Linguistics: Human Language
  Technologies}, pages 2523--2544.

\bibitem[{Raffel et~al.(2020)Raffel, Shazeer, Roberts, Lee, Narang, Matena,
  Zhou, Li, and Liu}]{raffel2020exploring}
Colin Raffel, Noam Shazeer, Adam Roberts, Katherine Lee, Sharan Narang, Michael
  Matena, Yanqi Zhou, Wei Li, and Peter~J Liu. 2020.
\newblock Exploring the limits of transfer learning with a unified text-to-text
  transformer.
\newblock \emph{Journal of Machine Learning Research}, 21:1--67.

\bibitem[{Rajpurkar et~al.(2016)Rajpurkar, Zhang, Lopyrev, and
  Liang}]{rajpurkar2016squad}
Pranav Rajpurkar, Jian Zhang, Konstantin Lopyrev, and Percy Liang. 2016.
\newblock Squad: 100, 000+ questions for machine comprehension of text.
\newblock In \emph{EMNLP}.

\bibitem[{Roberts et~al.(2020)Roberts, Raffel, and Shazeer}]{roberts2020much}
Adam Roberts, Colin Raffel, and Noam Shazeer. 2020.
\newblock How much knowledge can you pack into the parameters of a language
  model?
\newblock In \emph{Proceedings of the 2020 Conference on Empirical Methods in
  Natural Language Processing (EMNLP)}, pages 5418--5426.

\bibitem[{Su et~al.(2020{\natexlab{a}})Su, Xu, Dai, Ji, Yu, and
  Fung}]{su2020multi}
Dan Su, Yan Xu, Wenliang Dai, Ziwei Ji, Tiezheng Yu, and Pascale Fung.
  2020{\natexlab{a}}.
\newblock Multi-hop question generation with graph convolutional network.
\newblock In \emph{Proceedings of the 2020 Conference on Empirical Methods in
  Natural Language Processing: Findings}, pages 4636--4647.

\bibitem[{Su et~al.(2019)Su, Xu, Winata, Xu, Kim, Liu, and
  Fung}]{su-etal-2019-generalizing}
Dan Su, Yan Xu, Genta~Indra Winata, Peng Xu, Hyeondey Kim, Zihan Liu, and
  Pascale Fung. 2019.
\newblock \href {https://doi.org/10.18653/v1/D19-5827} {Generalizing question
  answering system with pre-trained language model fine-tuning}.
\newblock In \emph{Proceedings of the 2nd Workshop on Machine Reading for
  Question Answering}, pages 203--211, Hong Kong, China. Association for
  Computational Linguistics.

\bibitem[{Su et~al.(2020{\natexlab{b}})Su, Xu, Yu, Siddique, Barezi, and
  Fung}]{su2020caire}
Dan Su, Yan Xu, Tiezheng Yu, Farhad~Bin Siddique, Elham Barezi, and Pascale
  Fung. 2020{\natexlab{b}}.
\newblock Caire-covid: A question answering and query-focused multi-document
  summarization system for covid-19 scholarly information management.
\newblock In \emph{Proceedings of the 1st Workshop on NLP for COVID-19 (Part 2)
  at EMNLP 2020}.

\bibitem[{Su et~al.(2021)Su, Yu, and Fung}]{su-etal-2021-improve}
Dan Su, Tiezheng Yu, and Pascale Fung. 2021.
\newblock \href {https://doi.org/10.18653/v1/2021.findings-acl.275} {Improve
  query focused abstractive summarization by incorporating answer relevance}.
\newblock In \emph{Findings of the Association for Computational Linguistics:
  ACL-IJCNLP 2021}, pages 3124--3131, Online. Association for Computational
  Linguistics.

\bibitem[{Thorne et~al.(2018)Thorne, Vlachos, Christodoulopoulos, and
  Mittal}]{thorne2018fever}
James Thorne, Andreas Vlachos, Christos Christodoulopoulos, and Arpit Mittal.
  2018.
\newblock Fever: a large-scale dataset for fact extraction and verification.
\newblock In \emph{Proceedings of the 2018 Conference of the North American
  Chapter of the Association for Computational Linguistics: Human Language
  Technologies, Volume 1 (Long Papers)}, pages 809--819.

\bibitem[{Tombros and Sanderson(1998)}]{tombros1998advantages}
Anastasios Tombros and Mark Sanderson. 1998.
\newblock Advantages of query biased summaries in information retrieval.
\newblock In \emph{Proceedings of the 21st annual international ACM SIGIR
  conference on Research and development in information retrieval}, pages
  2--10.

\bibitem[{Trischler et~al.(2017)Trischler, Wang, Yuan, Harris, Sordoni,
  Bachman, and Suleman}]{trischler2017newsqa}
Adam Trischler, Tong Wang, Xingdi Yuan, Justin Harris, Alessandro Sordoni,
  Philip Bachman, and Kaheer Suleman. 2017.
\newblock Newsqa: A machine comprehension dataset.
\newblock In \emph{Proceedings of the 2nd Workshop on Representation Learning
  for NLP}, pages 191--200.

\bibitem[{Wang et~al.(2020)Wang, Cho, and Lewis}]{wang2020asking}
Alex Wang, Kyunghyun Cho, and Mike Lewis. 2020.
\newblock Asking and answering questions to evaluate the factual consistency of
  summaries.
\newblock \emph{Proceedings of the 58th Annual Meeting of the Association for
  Computational Linguistics}.

\bibitem[{Wu et~al.(2021)Wu, Galley, Brockett, Zhang, Gao, Quirk,
  Koncel-Kedziorski, Gao, Hajishirzi, Ostendorf et~al.}]{wu2021controllable}
Zeqiu Wu, Michel Galley, Chris Brockett, Yizhe Zhang, Xiang Gao, Chris Quirk,
  Rik Koncel-Kedziorski, Jianfeng Gao, Hannaneh Hajishirzi, Mari Ostendorf,
  et~al. 2021.
\newblock A controllable model of grounded response generation.
\newblock In \emph{Proceedings of the AAAI Conference on Artificial
  Intelligence}, volume~35, pages 14085--14093.

\bibitem[{Xu and Lapata(2020)}]{xu2020coarse}
Yumo Xu and Mirella Lapata. 2020.
\newblock Coarse-to-fine query focused multi-document summarization.
\newblock In \emph{Proceedings of the 2020 Conference on Empirical Methods in
  Natural Language Processing (EMNLP)}, pages 3632--3645.

\bibitem[{Yang et~al.(2018)Yang, Qi, Zhang, Bengio, Cohen, Salakhutdinov, and
  Manning}]{yang2018HotpotQA}
Zhilin Yang, Peng Qi, Saizheng Zhang, Yoshua Bengio, William Cohen, Ruslan
  Salakhutdinov, and Christopher~D Manning. 2018.
\newblock Hotpotqa: A dataset for diverse, explainable multi-hop question
  answering.
\newblock In \emph{Proceedings of the 2018 Conference on Empirical Methods in
  Natural Language Processing}, pages 2369--2380.

\bibitem[{Ye et~al.(2020)Ye, Li, Wang, Bolte, Ma, Yih, Ren, and
  Khabsa}]{ye2020studying}
Qinyuan Ye, Belinda~Z Li, Sinong Wang, Benjamin Bolte, Hao Ma, Wen-tau Yih,
  Xiang Ren, and Madian Khabsa. 2020.
\newblock Studying strategically: Learning to mask for closed-book qa.
\newblock \emph{arXiv preprint arXiv:2012.15856}.

\bibitem[{Young et~al.(2018)Young, Cambria, Chaturvedi, Zhou, Biswas, and
  Huang}]{young2018augmenting}
Tom Young, Erik Cambria, Iti Chaturvedi, Hao Zhou, Subham Biswas, and Minlie
  Huang. 2018.
\newblock Augmenting end-to-end dialogue systems with commonsense knowledge.
\newblock In \emph{Thirty-Second AAAI Conference on Artificial Intelligence}.

\bibitem[{Zhang et~al.(2021)Zhang, Sun, Gao, Fang, Brockett, Galley, Gao, and
  Dolan}]{zhang2021joint}
Yizhe Zhang, Siqi Sun, Xiang Gao, Yuwei Fang, Chris Brockett, Michel Galley,
  Jianfeng Gao, and Bill Dolan. 2021.
\newblock Joint retrieval and generation training for grounded text generation.
\newblock \emph{arXiv preprint arXiv:2105.06597}.

\bibitem[{Zhou et~al.(2018)Zhou, Prabhumoye, and Black}]{zhou2018dataset}
Kangyan Zhou, Shrimai Prabhumoye, and Alan~W Black. 2018.
\newblock A dataset for document grounded conversations.
\newblock In \emph{Proceedings of the 2018 Conference on Empirical Methods in
  Natural Language Processing}, pages 708--713.

\end{thebibliography}
\bibliographystyle{acl_natbib}


\appendix

\section{Implementation details}
\label{appendix:xxx}

We initialize our generation models with the pre-trained \textit{BART-large} models~\cite{lewis2020bart}, available
in the HuggingFace\footnote{\url{github.com/huggingface/transformers}} Transformers library. Our reader models was initiated from Span-BERT(base\&cased), from Facebook Github\footnote{\url{https://github.com/facebookresearch/SpanBERT}}, and further fine-tuned on MRQA datasets for 4 epochs, using the default fine-tuning configurations. Our RBG is trained using Adam~\cite{kingma2014adam} with a constant learning rate of 5e-5 and weight decay at 0.01. We train the model for 50k gradient steps, with a batch size of 4, using 8 Tesla V100 32Gb. We evaluate the models every 500 steps and select the best one on the validation set (1/8 of the evaluation set) based on the Rouge score. The maximum source document length is set to 300, and the target sequence length is set to 300. During inference, we use beam search with beam size of 4.

\section{Document retriever model details}
As the question/answers in LFQA may cover different domains and topics, we use a multi-task variant of DPR to guarantee the retrieval performance. The retriever is trained jointly on the union of all knowledge-intensive training data in KILT benchmark~\cite{petroni2021kilt}, including TrivaQA~\cite{Joshi_2017}, kwiatkowski2019naturaluestion~\cite{kwiatkowski2019natural}, HotpotQA~\cite{yang2018HotpotQA}, Fever~\cite{thorne2018fever}, zsRE~\cite{levy2017zero}, AY2, T-REx~\cite{elsahar2018t} and WoW~\cite{dinan2018wizard}.

\section{Human evaluation setup and analysis}
\label{appendix:human_eval}

\paragraph{Basic setup} As shown in Table~\ref{tab:human eval setup}, we sample 50 questions for each comparison and assign 3 annotators for each generated answer, which brings a workload of 450 judgments on model preference for each evaluation aspect. This process takes large amounts of energy and time considering the difficulty and challenges of factual-related annotation. We sample 10 questions from each of five development subsets corresponding to 5 levels of answer-passage overlap, which is a stratified sampling strategy. The answer position of each model is randomly shuffled to reduce the bias of position preference. 15 participants in our human evaluation are all researchers or students in computer science who speak and read English well.

\begin{table}[!ht]
\centering
\resizebox{0.45\textwidth}{!}
{
\begin{tabular}{c|ccc}
\hline
Comparison    & \#Questions & \#Annotators/answer  \\ \hline
RBG vs. FiD  &    50 &    3 \\
Reader analysis  &  50 &  3  \\
Pre-training analysis &  50 &  3 \\ \hline
\end{tabular}
}
\caption{Details of human evaluation for three comparisons.}
\label{tab:human eval setup}
\end{table}

\paragraph{Scoring setup} We ask each annotator to select a score from \{1,2,3\} for each generated answer in terms of three aspects: \textit{fluency}, \textit{relevance} and \textit{factual correctness}. During scoring, the annotators are asked to preserve the relative better-or-not relationship between two models as much as possible. In particular, for the metric of factual correctness, the annotators check the correctness of all factual statements involved in a generated answer by referring to Wikipedia~(EN), other web pages and the golden answer. The answer with significantly fewer factual errors will get a higher score on factual correctness. We show cases in Table~\ref{tab:human eval cases} to demonstrate how the annotator evaluate three aspects in our experiment.

\paragraph{Statistical analysis} We present the agreement among annotators on model preference in Table~\ref{tab:agreement} by calculating the Fleiss Kappa~\cite{fleiss_kappa} as the inter-rater consistency. The RBG vs. FiD comparison achieves better annotation agreement than other two ablation comparisons, maybe because RBG integrates both two of our contributions to improve the answer quality. In the comparison of RBG vs. FiD, annotators achieve a ``moderate agreement'' on the aspect of correctness and ``fair agreement'' on relevance~\cite{landis1977measurement}. Annotators achieve best agreements on fluency in all comparisons. It's more difficult to achieve a high degree of annotation consistency on factual correctness and relevance than fluency due to complicated facts involved in the generated answer. Therefore, we recommend taking preferred ratio as the core metric for factual-related evaluation following~\cite{krishna2021hurdles,nakano2021webgpt}. We also present score variance of four models involved in human evaluation in Table~\ref{tab:variance}. It's natural that the fluency score has the smallest variance while the difficult-to-annotate correctness has largest variance.

\begin{table}[!ht]
\centering
\resizebox{0.45\textwidth}{!}
{
\begin{tabular}{c|ccc}
\hline
Comparison    & fluency & relevance & correctness  \\ \hline
RBG vs. FiD  &    0.65 &    0.33 &    0.47 \\
Reader analysis  & 0.55 &    0.12 &    0.06  \\
Pre-training analysis&  0.62 &    0.16 &    0.08 \\ \hline
\end{tabular}
}
\caption{Agreement analysis for three comparisons in terms of three aspects. We use Fleiss Kappa~\cite{fleiss_kappa} to measure the agreement degree between annotators. The score range of [0,0.2] corresponds to slight agreement, [0.2,0.4 ] corresponds to fair agreement and [0.4,0.6] corresponds to moderate agreement~\cite{landis1977measurement}.}
\label{tab:agreement}
\end{table}

\begin{table}[!ht]
\centering
\resizebox{0.45\textwidth}{!}
{
\begin{tabular}{c|ccc}
\hline
Comparison    & fluency & relevance & correctness  \\ \hline
RBG  &    0.14 &    0.33 &    0.46 \\
RBG w/o reader & 0.22 &    0.34 &    0.48 \\
RBG w/o pre-training &  0.16 &    0.26 &    0.45 \\ 
RBG w/o reader+pre-training &  0.28 &    0.42 &    0.64 \\ \hline
\end{tabular}
}
\caption{Score variance of four models involved in human evaluation. }
\label{tab:variance}
\end{table}

\section{Number of retrieved documents on Generation Quality}

We also investigate the effects of number of retrieved documents $k$, on the answer generation quality. As we can see in Table~\ref{tab:results_k}, the generation quality in terms of ROUGE-L and F1, do not further improve as the number of $k$ increases, and the best performance are obtained when $k=10$ in our case.

\begin{table}[]\small
\centering
\begin{tabular}{c|cc}
\hline
ndocs & ROUGE-L & F1    \\ \hline
5     & 24.63   & 27.29 \\
10    & \textbf{24.72}   & \textbf{27.52} \\
20    & 24.39   & 26.68 \\
50    & 23.43   & 25.94 \\ \hline
\end{tabular}
\caption{Generation performance versus the number of retrieved documents of our model on MS MARCO~\cite{nguyen2016ms}.}
\label{tab:results_k}
\end{table}

\begin{table*}[htbp]\small
\begin{tabular}{|p{100pt}|p{250pt}|l|l|l|}
\hline
 Question & Generated answer & f & r & c \\ \hline
\multirow{2}{*}{effects of hyperthyroidism} &  Model1: Hyperthyroidism is a condition in which the thyroid gland produces too much thyroid hormone. The thyroid gland is a butterfly-shaped gland located in the neck. It is the only gland in the body that is responsible for the production of thyroid hormone, which helps regulate the body's metabolism. & 3 & 1 & 3 \\ \cline{2-5} 
 & Model2: Hyperthyroidism is a condition in which the thyroid gland produces too much thyroid hormone. This can cause the body to produce more thyroid hormone than it needs. This causes the thyroid to overproduce thyroid hormone, which can lead to a number of symptoms, including: 1  Weight gain. 2  Nausea. 3  Fatigue. 4  Vomiting. & 3 & 3 & 3 \\ \hline
\multirow{2}{*}{where is cabo de hornos} &  Model1: Cabo de Hornos is the southern tip of South America's Cape Horn Archipelago. It is the \hlgreen{largest island in the Pacific Ocean} and is the \hlgreen{only island in South America with a population of more than 2,000,000 inhabitants}. The island is also the \hlgreen{world's second-largest island} in terms of land area, with a total area of \hlgreen{1,826 square miles (4,926 square kilometers)} . & 3 & 2 & 1 \\ \cline{2-5} 
 &  Model2: Cabo de Hornos, Chile. Cabo de Hanoque, also known as Cape Horn, is the southern tip of South America. It is located in the south of Tierra del Fuego archipelago, in the AntÃ¡rtica Province, Magallanes Region. The municipality of Cabo De Hornos is the capital of the province of Puerto Williams. & 3 & 3 & 3 \\ \hline
\end{tabular}
\caption{Cherry-picked scoring cases by annotators in our human evaluation. f, r, and c respect fluency, relevance and correctness respectively. We use \hlgreen{green} to mark out factual incorrect snippets the annotator finds. The first case shows how annotators distinguish different levels of relevance and the second case shows that of factual correctness.}
\label{tab:human eval cases}
\end{table*}

\begin{table*}[!th]
\small
\centering
\begin{adjustbox}{width={0.90\textwidth},totalheight={\textheight},keepaspectratio}
\begin{tabular}{p{2\columnwidth}}
\hline
\textbf{Question (from NQ):} what are the ranks in the us navy? \\
\hline
\textbf{Gold Answer:} ["E-8s senior chief petty officer", "E-9s master chief petty officer"] \\
\textbf{BART:}  (?) Navy ranks in the U.S. Navy. The Navy ranks are as follows: 1  Master Chief Petty Officer (MOS) a non-commissioned officer in the Navy. 2  Master Sergeant (MS), a Master Sergeant in the Marine Corps. 3  Sergeant Major (SSM), a Sergeant Major in the Air Force. \\
\textbf{DPR-BART:}  (?) The U.S. Navy is a branch of the United States Armed Forces. The Navy is made up of the Navy, Marine Corps, Air Force, Coast Guard, and Coast Guard.\\
\textbf{FiD:}  (*) The U.S. Navy is a branch of the United States Armed Forces. The rank system of the Navy consists of three categories: junior officers, senior officers, and flag officers. Junior officers are those officers in pay grades O-1 to O-4, \hlgreen{while senior officers are the ones with pay grades of O-5 and O-6. Senior officers have pay grades ranging from O-7 to E-9}. \hlgreen{Flag officers have a pay grade ranging from E-5 to E9}. \\
\textbf{RBG}(ours): The U.S. Navy ranks are divided into three categories: junior officers, senior officers, and flag officers. Junior officers are officers in pay grades O-1 to O-4, while senior officers are those in pay grades O-5 and O-6. Flag officers serve either as a line officer or as a staff corps officer. \\
\hline
\hline
\textbf{Question (from HotpotQA):} Which  French ace pilot and adventurer fly L'Oiseau Blanc?  \\
\hline
\textbf{Gold Answer}:["Charles Eugène"] \\
\textbf{BART:} (?) L'oiseau Blanc is a French aviator and aviator who flew the L'Oiseau blanc. He is best known for being the first aviator to fly it.. \\
\textbf{DPR-BART:} (?) L'oiseau Blanc l'oiseau blanc (French: The White Bird) was a French Levasseur Pl.8 biplane that disappeared in the early morning hours of April 1, 2014. The plane was a Levassellseur PL.8 Biplane that vanished in the morning of April 2, 2014, after being piloted by an adventurer pilot and adventurer pilot..\\
\textbf{FiD:} (*) On May 8, 1927, Charles Nungesser and Francois Coli took off from Paris on the L'Oiseau Blanc, a Levasseur PL.8 biplane \hlgreen{that had been designed by Charles Nongesser}, a French ace pilot and aviator, to fly from Paris to New York City in the Orteig Prize. \\
\textbf{RBG(ours)}:  On May 8, 1927, Charles Nungesser and Francois Coli boarded L'Oiseau blanc, a 450-hp Lorraine-powered Levasseur biplane designed to compete for the Orteig Prize. They took off from Paris on 8 May 1927 and were last seen over Ireland. Less than two weeks later, Charles Lindbergh successfully made the New York-Paris journey and claimed the prize in the Spirit of St. Louis.\\
\hline

\end{tabular}
 \end{adjustbox}

\caption{Examples of the zero-shot long-form answers on the NQ~\cite{kwiatkowski2019natural} and HotpotQA ~\cite{yang2018HotpotQA} datasets. RBG model generates more fluent, specific and factually accurate responses. ‘?’ indicates factually incorrect/irrelevant  responses; * indicates partially correct responses. We use the \hlgreen{green} to mark out the unfaithful snippets.}
\label{tab:examples}
\end{table*}

\section{Case of zero-shot experiments on NQ and HotpotQA}
\label{appendix:results}
In addition to automatic evaluation on the two datasets, we also show concrete examples to give a better understanding. As we can see in Table~\ref{tab:examples}, ~\textbf{BART} tends to generate unfaithful answers as it lacks of external knowledge related to the specific question, and \textbf{DPR + BART} produces somewhat faithful but irrelevant answers, while ~\textbf{FiD} performs much better but still with partially unfaithful generations. In comparison, our ~\textbf{RBG} can generate more fluent, specific and factually accurate responses.


\end{document}